# Poster: Camera Tampering Detection for Outdoor IoT Systems

Shadi Attarha, Kanaga Shanmugi, Anna Förster
Dept. Sustainable Communication Networks
University of Bremen, Germany
Corresponding author: sattarha@uni-bremen.de

## Abstract

Recently, the use of smart cameras in outdoor settings has grown to improve surveillance and security. Nonetheless, these systems are susceptible to tampering, whether from deliberate vandalism or harsh environmental conditions, which can undermine their monitoring effectiveness. In this context, detecting camera tampering is more challenging when a camera is capturing still images rather than video as there is no sequence of continuous frames over time. In this study, we propose two approaches for detecting tampered images: a rule-based method and a deep-learning-based method. The aim is to evaluate how each method performs in terms of accuracy, computational demands, and the data required for training when applied to real-world scenarios. Our results show that the deep-learning model provides higher accuracy, while the rule-based method is more appropriate for scenarios where resources are limited and a prolonged calibration phase is impractical. We also offer publicly available datasets with normal, blurred, and rotated images to support the development and evaluation of camera tampering detection methods, addressing the need for such resources.

## CCS Concepts

• **Computer systems organization** → **Reliability**; **Sensors and actuators**; • **Computing methodologies** → **Machine learning approaches**.

## Keywords

Detection algorithms, Cameras, Surveillance, Reliability, Deep learning, Rule-based systems, Image processing

## 1 Introduction and Motivation

Nowadays, surveillance cameras are widely used for object recognition or environmental monitoring. However, their performance can degrade when deployed in unattended locations due to factors such as deliberate sabotage or environmental conditions like storms. To ensure these systems remain accurate, it is vital to have a detection system that can identify tampered images while requiring minimal resources [1].

Notable works have been done to address tampering challenges in video applications. For instance, Pan et al. [6] use Multi-Layer Perceptron (MLP) and ResNet models specifically to detect blurred videos. Other studies [4, 5] leverage Convolutional Neural Networks (CNNs) to detect tampering in videos. Although the mentioned approaches aim to identify alterations in visual content, video tampering detection benefits from continuous frames that provide temporal context, unlike static image tampering detection. Furthermore, these techniques require substantial computational power to run complex deep learning models, which poses a problem when trying to deploy them on resource-limited devices like the ESP32 or even Raspberry Pi 2.

Our motivation stems from an animal detection system that captures images periodically or upon detecting movement [3]. Hence, our goal is to identify and diagnose camera tampering issues, including blurred images, rotated images, and obstructed lenses. To achieve this, we explore two distinct approaches: a lightweight CNN-based model and a rule-based method. We evaluate these methods in terms of performance metrics and resource requirements, providing insights into their advantages and limitations when deployed on resource-constrained edge devices. We also provide labeled datasets with a diverse set of images to support the development and evaluation of tampering detection algorithms.

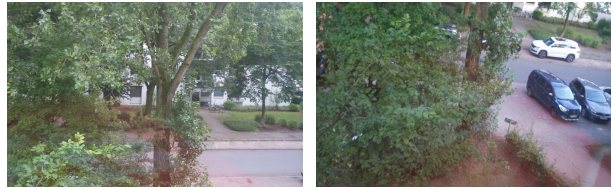

Figure 1: Normal image (left), Rotated image (right)

## 2 Hardware Setup and Methods

The hardware setup includes a Raspberry Pi 2 Model B, Camera Module 3, and an SD card for storage. Over 10 days, 3600 images were collected: 1800 normal, 600 blurred, 600 rotated, and 600 with obstructed lenses. Two sample images are shown in Figure 1, and the datasets are available online[1].

### 2.1 Rule-based method

The proposed rule-based method relies solely on normal images to define a baseline that represents expected normal patterns in various environmental conditions. This baseline then serves as a reference for comparison, helping to identify deviations in new images that might indicate tampering. In this study, eight normal images, representing different times of day (i.e. different lighting conditions during night, morning, noon, and evening), and different weather conditions (i.e. sunny, windy, cloudy, and rainy), are used as references and to simplify the analysis, the color images (RGB) are converted to grayscale.

In the first step, the evaluation focuses on identifying if an image is blurred or obstructed. This is because the statistical properties, like mean and variance, show more pronounced differences between normal and blurred/obstructed images, whereas the differences between normal and rotated images might be less significant. Hence, the ORB method (Oriented FAST and Rotated BRIEF) [7] is used to identify keypoints in both test and reference images. These features are then compared using the BFMatcher (Brute Force Matcher) algorithm which is a part of the OpenCV library [2]. This comparison results in a number of matches between the test and a reference image. The image is promptly flagged as abnormal if the number of good matches is below the threshold. This threshold is established during the calibration phase by comparing differences between normal images, ensuring it accurately reflects the expected variation in normal conditions. Further checks are then conducted to identify whether the tampered image belongs to the blurred or no-image category. To detect blur, we use the Laplacian operator. If the image's sharpness is too low, it is classified as blurred. If the image is not blurred, we calculate the pixel intensity standard deviation. A low value indicates an empty or uniform image, classifying it as a no-image case [4]. On the other hand, if the number of good matches is above the threshold, we estimate the image's rotation using the homography matrix [7]. If the rotation angle exceeds 50°, the image is classified as rotated. Otherwise, the image will be classified as normal.

---

[1] https://nc.uni-bremen.de/index.php/s/GCsBcswrQf9kBXC

## 2.2 Deep Learning based method

The CNN-based model in this research is designed to classify images as normal or abnormal and is implemented using TensorFlow and Keras. The architecture begins with a sequence of convolutional layers: the first layer has 32 filters, followed by layers with 64 and 128 filters. Each convolutional layer is paired with ReLU activation and max-pooling to reduce dimensionality. After these layers, the output is flattened and passed through a fully connected layer with 128 units. The final layer, with a single unit and a sigmoid activation function, is tailored for binary classification. To train the model, we use a robust dataset of 2400 images, equally divided between normal and abnormal categories. Normal images are captured under various weather conditions to incorporate diverse lighting and environmental factors, while the 1200 abnormal images reflect common surveillance issues such as blurred, and rotated images. This diverse dataset helps the CNN learn effectively and generalize well to new data.

## 2.3 Comparison

We evaluated the performance of the rule-based method and the CNN-based model using a balanced dataset comprising 1200 images, with a 50% ratio of abnormal images to avoid bias and ensure reliable results. Table 1 presents the performance metrics for each approach. The results demonstrate that the CNN model outperforms the rule-based model in correctly classifying images. However, the CNN model's dependency on a comprehensive training dataset is a notable limitation. If the camera is repositioned, a new dataset comprising all possible faulty images needs to be collected, which increases both resource demands and time. Additionally, the size of the CNN model presents a challenge when deploying it on resource-constrained edge devices such as microcontrollers or Raspberry Pi 2, where memory limitations may prevent deployment. In contrast, the rule-based method, despite its lower performance, offers advantages in resource-limited environments. With a model size of 8.77 KB and minimal training requirements, it is ideal for scenarios where large training datasets are impractical to collect or edge devices have limited resources. Furthermore, as it does not depend on tampered or abnormal images, it presents a feasible solution in situations where collecting such images is challenging, time-consuming, and costly. It is important to mention that increasing the number of reference images would enhance the performance of the method. As such, our comparison suggests that the rule-based method remains a viable alternative to CNN in resource-constrained environments.

Table 1: Performance evaluation

| Metrics | Rule-based model | CNN model |
|---|---|---|
| Accuracy | 0.9050 | 0.9975 |
| Precision | 0.9036 | 0.9983 |
| F1-Score | 0.9051 | 0.9974 |
| Recall | 0.9066 | 0.9966 |
| Number of training/reference Samples | 8 | 2400 |
| Number of abnormal samples in training set | 0 | 1200 |
| Size of the model | 8.77 KB | 37.8 MB |
| Average processing time in testing phases (Per image) | 0.07 sec | 0.02 sec |

## 3 Conclusion

This paper evaluates two camera tampering detection methods: a rule-based approach and a CNN-based model. Labeled datasets were compiled to provide diverse images, supporting the development and testing of detection algorithms. Results show that while the rule-based method proves to be more resource-efficient, the CNN model demonstrates superior accuracy and robustness. The choice between these methods depends on deployment needs, such as available computational resources and the desired accuracy level. Future work will focus on developing a hybrid system that integrates rule-based and lightweight CNN approaches to balance performance and resource usage. Additionally, we plan to evaluate these models using datasets collected from multiple locations.